\documentclass[letterpaper]{article} 
\usepackage{aaai2026}  
\nocopyright
\usepackage{times}  
\usepackage{helvet}  
\usepackage{courier}  
\usepackage[hyphens]{url}  
\usepackage{graphicx} 
\usepackage{natbib}  
\usepackage{caption} 
\usepackage{tabularx}     
\usepackage{booktabs}     
\usepackage{multirow}     
\usepackage{array}        
\usepackage{amsmath}
\usepackage{amsfonts} 
\usepackage{xcolor}
\usepackage{algorithm}
\usepackage{algorithmic}
\usepackage{makecell}

\usepackage{newfloat}
\usepackage{listings}
\DeclareCaptionStyle{ruled}{labelfont=normalfont,labelsep=colon,strut=off} 
\floatstyle{ruled}
\newfloat{listing}{tb}{lst}{}
\floatname{listing}{Listing}
\title{Multi-modal Reasoning with LLMs for Visual Semantic Arithmetic}
\author{
    Chuou Xu,
    Liya Ji,
    Qifeng Chen\textsuperscript{\dag}
}

\affiliations{
    The Hong Kong University of Science and Technology\\
    cxubw@connect.ust.hk, lji@connect.ust.hk, cqf@ust.hk\\
    \textsuperscript{\dag}Corresponding author
}

\usepackage{bibentry}

\begin{document}

\maketitle
\begin{abstract}

Reinforcement learning (RL) as post-training is crucial for enhancing the reasoning ability of Large Language Models (LLMs) in coding and math. 
However, their capacity for \emph{visual semantic arithmetic}—inferring relationships from images—remains underexplored. 
The classic text analogy ``king''--``man''+``woman'' = ``queen'' illustrates relational reasoning, yet replacing text with images of ``king'' and ``man'' significantly reduces performance due to the need for commonsense knowledge and the extraction of concise concepts over irrelevant visual details. Such capability is vital for service and domestic robotics in unstructured environments, where robots must infer semantic relationships between objects, agents, and actions.
In a kitchen, recognizing from images the triple ``powder''--``cake'' = ``is made of'' demonstrates grounding of symbolic relations in perception, enabling tool substitution (e.g., spatula for spoon), task generalization (recipe adaptation), and improved semantic reasoning. Prior work approaches semantic arithmetic by decoding image features after vector arithmetic, but suffers from modality gaps and lacks systematic evaluation. 
In this paper, we formulate two novel tasks: (1) two-term subtraction and (2) three-term operations, and construct the Image-Relation-Pair Dataset (IRPD) for benchmarking. 
We further propose \textbf{Semantic Arithmetic Reinforcement Fine-Tuning (SAri-RFT)}, which post-trains Large Vision-Language Models (LVLMs) using a verifiable function and \emph{Group Relative Policy Optimization} (GRPO). 
Our method achieves state-of-the-art results on IRPD and the real-world Visual7w-telling dataset~\cite{zhu2016visual7w}. 
By equipping LVLMs with robust cross-modal relational reasoning, this work advances domestic robots' ability to ground symbolic reasoning in perception, enhancing decision-making, tool adaptability, and human–robot interaction in complex environments.
Datasets and source code are provided in the supplementary material.

\end{abstract}

\begin{links}
    \link{Project Page}{https://github.com/xcooool/vis_arithmetic}
\end{links}
\section{Introduction}

Model reasoning ability becomes increasingly important in solving complex tasks like mathematical problems.
Current Large Language Models (LLMs) like GPT-4~\cite{achiam2023gpt} or DeepSeek-R1~\cite{guo2025deepseek} shows a strong reasoning capability in math and coding problems.
However, the performance of those models drops in solving visual semantic arithmetic tasks due to the high requirements of multi-modal understanding and prior knowledge of the world.
A famous analogy can be written as the following arithmetic equation: ``king'' - ``man'' + ``woman'' = ``queen.'' 
We can use Large Vision-Language Models (LVLMs) to solve this problem easily with the text input only, but the reasoning ability decreases when facing the visual modality.



Solving visual semantic arithmetic tasks requires models to integrate high-level conceptual understanding with selective attention to relevant semantic cues.
The first challenge is that the model needs to have common sense before solving semantic arithmetic problems. 
For example, ``a swimming pool'' - ``parties'' = ``a saddle'' - ``a horse,'' with both relations being ``used for.''
The second challenge in visual semantic arithmetic is that models must abstract the overall meaning of each image input into a concise phrase, rather than focusing on irrelevant visual details, particularly when the image depicts a complex scene with both foreground and background~\cite{huang2019attention, zhang2021exploring}.





ZeroCap~\cite{tewel2022zerocap} is the first work to demonstrate the ability to solve visual semantic arithmetic tasks by performing the arithmetic operation on image embedding. However, it is limited by the instability and vulnerable to the modality gap.
To narrow the modality gap, recent research~\cite{zhu2023languagebind, girdhar2023imagebind} aims to map different modalities into a joint embedding space.
However, they do not guarantee that the vector relations among concepts are preserved in a unified embedding space, which is vital for solving visual semantic arithmetic tasks.
To quantify the models' performance in semantic arithmetic problems, ~\cite{couairon2022embedding} proposed the SIMAT dataset, which uses triplets in the format $(subject, relation, object)$ for image retrieval. However, its fixed vocabulary of only 10 subjects and 20 object items limits scalability and the ability to address a broad range of semantic arithmetic problems without a general LLM.

This paper proposes a novel framework for visual semantic arithmetic reasoning, combining a new task formulation, a multimodal dataset, and an effective fine-tuning method. Firstly, we define a formulation to represent the metadata and design two semantic arithmetic tasks: a two-term subtraction task, which consists of multiple-choice questions to select the relation between two items, and a three-term operations task, which is similar to analogy problems given three items.
Inspired by ConceptNet~\cite{speer2017conceptnet}, we construct a comprehensive dataset comprising 18 relations and over 1500 subject-object pairs, each represented in both text and corresponding image formats, according to the formulation.
Moreover, in contrast to the embedding arithmetic method used by Zerocap~\cite{tewel2022zerocap}, we propose Semantic Arithmetic Reinforcement Fine-Tuning (SAri-RFT), which post-trains LVLMs to solve visual semantic arithmetic problems by incorporating Reinforcement Learning with Verifiable Rewards (RLVR)~\cite{liu2025visual} with a newly designed verifiable reward function. 
The Large Visual Language Model enables our method to have commonsense knowledge and allows understanding the core semantic content of an image, rather than being influenced by unnecessary visual elements.
SAri-RFT achieves a huge leap in two visual semantic arithmetic tasks compared to Supervised Fine-Tuning (SFT) and the embedding arithmetic method. 

Our contributions can be summarized as follows:
\begin{itemize}
  \item We formulate two semantic arithmetic tasks, a two-term subtraction task and a three-term operations task, and construct an Image-Relation-Pair Dataset(IRPD) based on the formulation, which consists of 18 relations and over 1500 high-quality subject-object pairs in both text and image modality.
 \item We propose Semantic Arithmetic Reinforcement Fine-Tuning (SAri-RFT), post-training an LVLM with a newly designed soft verifiable reward function via Group Relative Policy Optimization (GRPO).
 \item We conduct extensive experiments on the IRPD dataset and demonstrate the state-of-the-art performance with more than $50\%$ improvement compared to other methods. We also show that the effectiveness of SAri-RFT on a real-world downstream task, Visual7w-telling~\cite{zhu2016visual7w}, outperforms SFT and the embedding arithmetic method.
\end{itemize}

\section{Background}

\subsection{Embedding Arithmetic}
Existing approaches for semantic arithmetic tasks can be broadly categorized into two types: embedding arithmetic methods, which leverage vector operations on embeddings but are often unstable and suffer from modality gaps, and LVLMs reasoning, which frames the task as a Visual Question Answering (VQA) problem.
Many studies have demonstrated the arithmetic properties of text embedding using various models and metrics, including Word2Vec~\cite{church2017word2vec}, GloVe~\cite{pennington2014glove}, fastText~\cite{joulin2016fasttext}, and BERT~\cite{koroteev2021bert}. With the development of multi-modal alignment and integration, similar arithmetic properties are now being observed in image embeddings~\cite{girdhar2023imagebind, zhu2023languagebind, li2020oscar,anwaar2021compositional,engilberge2018finding}. 
The SIMAT dataset~\cite{couairon2022embedding} has demonstrated the feasibility of image embedding arithmetic for visual semantic arithmetic, while consisting of only 10 predefined subject words and 20 object words, posing challenges for extending the approach to broader application domains.

ZeroCap\cite{tewel2022zerocap} tries to solve these problems by utilizing CLIP~\cite{radford2021learning}'s image encoder to perform arithmetic operations on image embeddings, enabling text generation with the integration of GPT-2~\cite{radford2019language}. However, multi-model embedding arithmetic is vulnerable to the modality gap. Moreover, generating each word for ZeroCap requires five rounds of updates in GPT-2’s attention matrix~\cite{guo2022attention}, which is extremely time-consuming and unstable.
To narrow the modality gap, ImageBind~\cite{girdhar2023imagebind} and LanguageBind~\cite{zhu2023languagebind} have been proposed to learn a joint embedding across modalities, showing strong few-shot abilities in recognition tasks. However, they focus only on cross-modal alignment while ignoring the vector relations within each modality. 
They try to make the text and image embeddings of the same item (e.g., "egg") as close as possible. However, this doesn’t ensure that the differences between pairs of items (like "egg" vs. "chicken") in text and image embeddings point in similar directions. This suggests that aligning individual embeddings doesn’t guarantee that relationships between items are preserved across modalities.
\begin{figure}[t]
    \centering
\includegraphics[width=.48\textwidth]{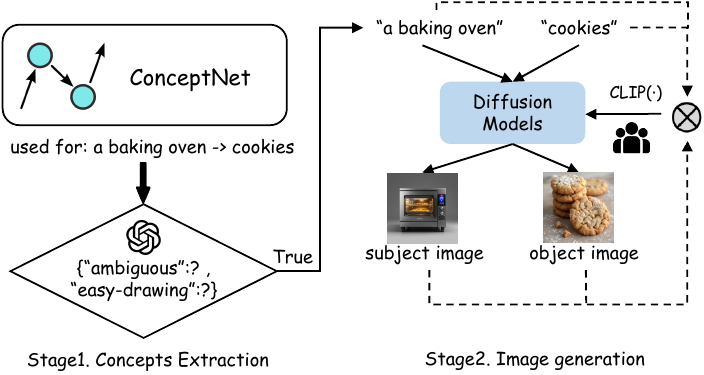}
    \caption{The pipeline of IRPD dataset generation. Firstly, we extract triplets ($relation: subject \rightarrow object$) from ConceptNet~\cite{speer2017conceptnet}. Secondly, we generate the images using text-to-image diffusion models, such as Flux~\cite{flux2024}. Moreover, LLMs validation, a CLIP-based selection criterion, and human evaluation are applied to ensure the quality of subject-object pairs in the text and image modality, respectively.}
    \label{fig:dataset_gen}
\end{figure}
\subsection{LVLMs Reasoning}
If we treat visual semantic arithmetic as a VQA~\cite{li2023comprehensive} task, LVLMs have shown great potential for these visual reasoning tasks, such as CLEVR~\cite{johnson2017clevr}, FigureQA~\cite{kahou2017figureqa}, and GuessWhat?!~\cite{de2017guesswhat}.
Furthermore, inspired by DeepSeek-R1~\cite{guo2025deepseek}, current LVLMs prefer post-training to enhance reasoning ability with various methods, including Retrieval-Augmented Generation (RAG)~\cite{lewis2020retrieval}, Chain-of-Thought (CoT) reasoning~\cite{wei2022chain}, Supervised Fine-Tuning (SFT)~\cite{zhang2023instruction}, and Reinforcement Learning with Human Feedback (RLHF)~\cite{christiano2017deep}. 

\begin{figure*}[t]
    \centering
    \includegraphics[width=.9\textwidth]{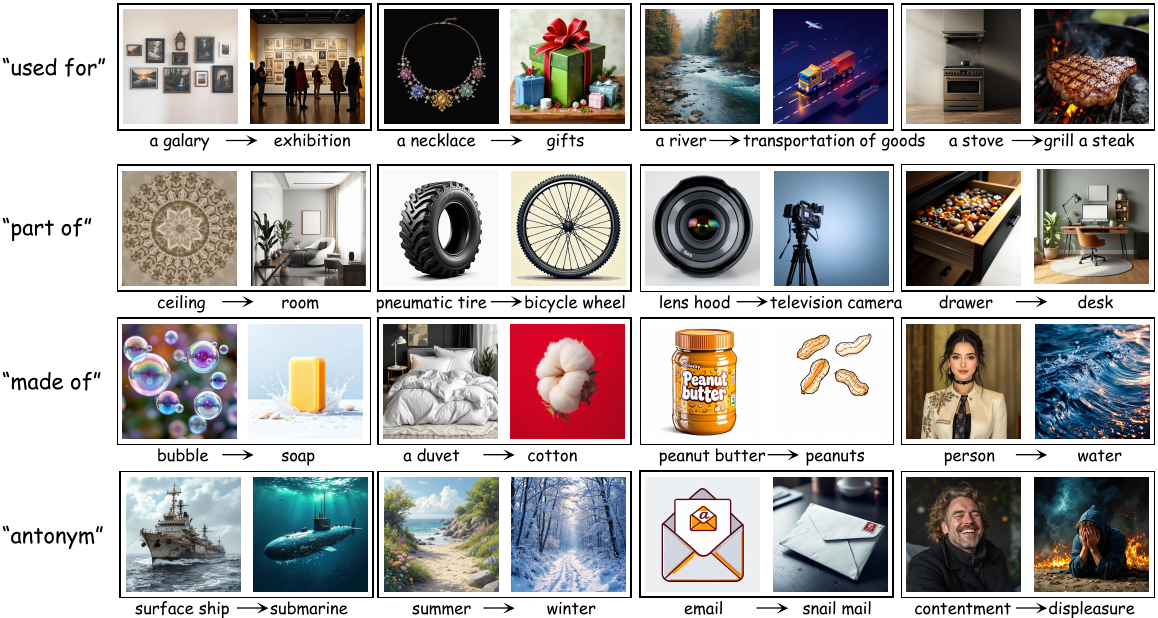}
    \caption{Examples of the IRPD Dataset. We show four relations in this figure, including ``used for,'' ``part of,'' ``made of,'' and ``antonym.'' Each line illustrates four subject-object pairs corresponding to each relation in text and image modality.}
    \label{fig:flux_dataset}
\end{figure*}
DeepSeek-R1~\cite{guo2025deepseek} has demonstrated the huge impact of Reinforcement Learning(RL) in the post-training stage using GRPO with a verifiable reward function. 
Unlike Supervised Finetuning (SFT)~\cite{zhang2023instruction}, which requires a large number of annotated data, and Reinforcement Learning with Human Feedback (RLHF)~\cite{christiano2017deep}, which is hard to collect the data due to its reliance on human preference annotations, RLVR depends on automatic evaluations based on fixed rules with limited hundreds of data, reducing biases in human judgment and increasing the robustness and scability of post-training stage.
Specifically, for each question input $q$, the old policy model $\pi_{\theta_{old}}$ will sample a group of $n$ responses $\{o_i\}_{i=1}^n$, each including the thinking process and the final answer $a_i$. Afterwards, the rewards of these responses will be normalized to calculate the group relative advantages $\hat{A}_{i,t}$:
\begin{align}
    & \hat{A}_{i,t} = \frac{R(o_i) - mean(\{R(o_i)\}_{i=1}^n)}{std(\{R(o_i)\}_{i=1}^n)}
\end{align}
where $R$ denotes a verifiable reward function.
The goal is to maximize the following clipped objective~\cite{guo2025deepseek}:
\begin{align}
    J_{\text{GRPO}}(\theta) &= \mathbb{E}_{\{o_i\}_{i=1}^n \sim \pi_{\theta_{\text{old}}}(O | q)} \notag \\
    &\quad \frac{1}{n} \sum_{i=1}^n \frac{1}{|o_i|} \sum_{t=1}^{|o_i|} \left\{ \min \left[ \frac{\pi_{\theta}(o_{i,t} | q, o_{i,<t})}{\pi_{\theta_{\text{old}}}(o_{i,t} | q, o_{i,<t})} \hat{A}_{i,t}, \right. \right. \notag \\
    &\quad \left. \left. \text{clip} \left( \frac{\pi_{\theta}(o_{i,t} | q, o_{i,<t})}{\pi_{\theta_{\text{old}}}(o_{i,t} | q, o_{i,<t})}, 1 - \epsilon, 1 + \epsilon \right) \hat{A}_{i,t} \right] \right. \notag \\
    &\quad \left. - \mu D_{\text{KL}}[\pi_{\theta} || \pi_{\text{ref}}] \right\}
\end{align}
where $\pi_\theta$ denotes the current policy model, $\pi_{ref}$ denotes the reference frozen model, and $\epsilon, \mu$ are hyper-parameters.
However, in previous studies~\cite{gandhi2024stream, lambert2024t, ma2025s, liu2025visual}, the verifiable reward function for GRPO is mainly a solid binary function with 0 or 1, which has a lot of limitations for open-ended questions like analogy reasoning problems.
\section{Method}
\subsection{Overview} \label{Overview}

\begin{figure*}[t]
    \centering
    \includegraphics[width=\textwidth]{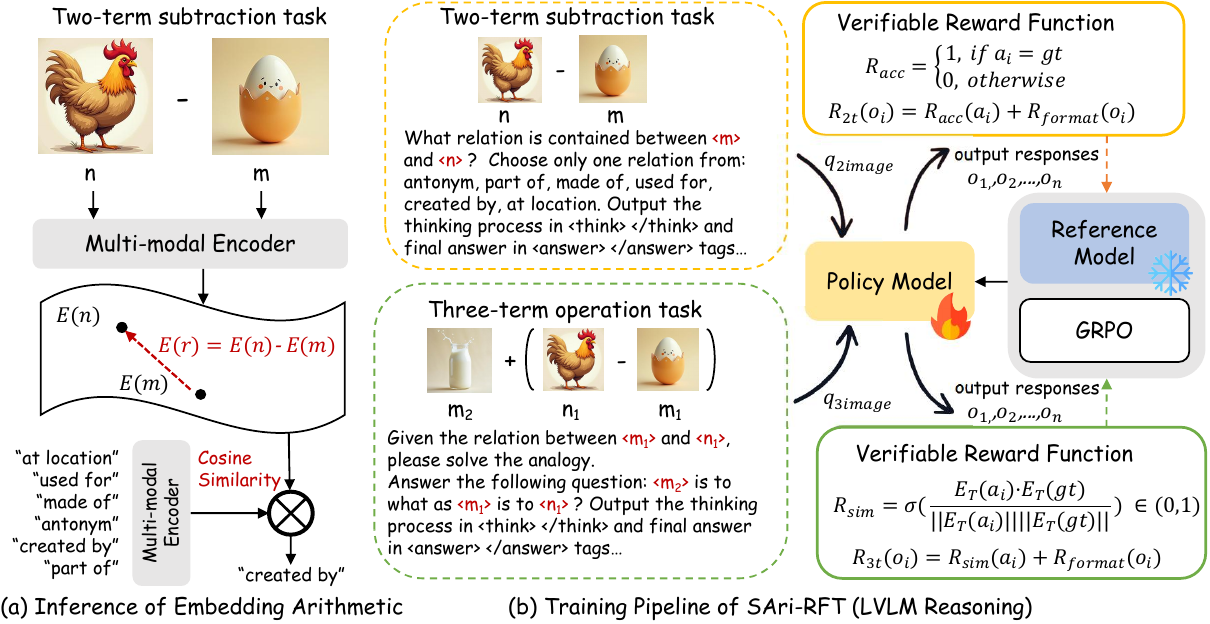}
    \caption{
    (a) Inference of embedding arithmetic for the two-term subtraction task with image-based subject-object pair input $q_\text{2image}$, where m, n represent subject and object, respectively. $E$ refers to the multi-modal encoder.
    (b) Training pipeline of SAri-RFT for question input $q_\text{2image}$ and $q_\text{3image}$, where $E_{T}$ refers to text encoder. For each question input, the LVLM policy model first samples a group of $n$ responses $\{o_i\}_{o=1}^n$, including the thinking process and the final answer $a_i$. Secondly, GRPO will calculate the group relative advantages to optimize the model with a dual verifiable reward function.
    }
    \label{fig:pipeline}
\end{figure*}

In this paper, our work is the first to formally define two visual semantic arithmetic tasks, as two-term subtraction task and the three-term operation task. 
Moreover, we construct a comprehensive IRPD dataset for systematic evaluation and propose SAri-RFT in Figure~\ref{fig:pipeline} (b), which leverages an LVLM post-trained with RLVR to address visual semantic arithmetic as a VQA task, in contrast to ZeroCap in Figure~\ref{fig:pipeline} (a), which performs visual semantic arithmetic directly in the embedding space.

\paragraph{Two-term Subtraction Task} 

Firstly, we define a formulation as $relation: subject \rightarrow object$, where both the subject and object can be either text or image, while the relation is always represented with text. For example, ``created by: egg → chicken.'' 
Secondly, after mapping the above triplet into the embedding space, the relation is projected as a vector from subject to object, so the formulation can be rewritten as $E(relation) = E(object)-E(subject)$, where E denotes the multi-modal encoder. 
Furthermore, to simplify the equation, we denote the two-term subtraction task as $object-subject=relation$, which requires the models to infer the relation with subject-object pair input.
Since it is difficult for models to generate the text relation accurately, the two-term subtraction task is designed as a multiple-choice question requiring the model to select the correct relation. 

\paragraph{Three-term Operation Task}
Firstly, we extend the two-term subtraction task into the formulation of $E(object_1) - E(subject_1)=E(object_2)-E(subject_2)$, where the relations between the two subject-object pairs are identical, resembling analogy problems.
To simplify the equation, we denote the three-term operation task as $object_1 - subject_1 + subject_2 = object_2$, which requires the models to generate the text response representing $object_2$ with three-term input.
For example, ``author - book + child = birth.''
Since the relations we select follow a many-to-one mapping, we designate the object as our task's ground truth instead of the subject.

\subsection{IRPD Dataset}
\label{sec: dataset}
\subsubsection{Text-Based Relation Dataset}

Following the formulation $relation: subject \rightarrow object$ defined in the two-term subtraction task, we leverage a text semantic multilingual network, ConceptNet~\cite{speer2017conceptnet}, to retrieve the text triplets.
Firstly, from the 32 relations available in ConceptNet, we select 18 for our dataset, excluding abstract relations such as ``form of,'' ``motivated by'' and ``symbol of.''
Secondly, for each relation, we retrieve all text-based subject-object pairs from ConceptNet, where the subject or object is typically a single word or a short phrase of up to three words instead of a sentence.

Moreover, each pair will be validated using GPT-4~\cite{achiam2023gpt} to ensure they are visually easy to represent.
Pairs containing verbs or any ambiguity will be filtered out, thereby significantly improving the quality of the selected text-based subject-object pairs.
Ultimately, the text-based relation dataset has been successfully constructed. The pipeline is shown in Figure~\ref{fig:dataset_gen}.


\subsubsection{Image-Based Relation Dataset}

After obtaining text-based subject-object pairs, we employ Text-to-Image Models to generate corresponding image-based subject-object pairs. Instead of using Stable Diffusion ~\cite{podell2023sdxl}, which tends to produce blurry images, distort human figures, and suffer from instability, we utilize FLUX.1-dev ~\cite{flux2024}, a 12-billion-parameter model known for its superior image quality. For each generated image, we filter the images by calculating the CLIP-score ~\cite{radford2021learning} to evaluate the semantic alignment between the image and the corresponding subject or object text, followed by manual verification to ensure visual consistency and correctness.
Figure~\ref{fig:flux_dataset} shows examples of the IRPD dataset.

\subsection{Semantic Arithmetic Reinforcement Fine-Tuning}
Due to the modality gap, instability, and extensive time consumption caused by the embedding arithmetic method used by ZeroCap, we adopt LVLMs to address two visual semantic arithmetic tasks as a VQA task, which leverages the model's multi-modal reasoning capabilities.
Inspired by DeepSeek-R1~\cite{guo2025deepseek}, reinforcement learning with verifiable rewards (RLVR) has been proven to dramatically improve the reasoning ability in mathematics and coding tasks with structured reference answers.
For two visual semantic arithmetic tasks, we design different verifiable reward functions such as $R_{2t}, R_{3t}$ to replace $R$ in equation (1). 

\subsubsection{Two-term subtraction task implementation} \label{subtraction method}
For the two-term subtraction task, we design it as multiple-choice questions like a classification problem, where the model is required to choose the right answer with a thinking process before answering. 
For the formulation $object-subject=relation$, we design the prompt below to solve this task as a VQA task, where both $\langle subject \rangle$ and $\langle object \rangle$ can be replaced by text or image:
\begin{itemize}
    \item What relation is contained between $\langle subject \rangle$ and $\langle object \rangle$? Choose only one phrase from: $r_1,r_2,...,r_p$
\end{itemize}
where $r$ denotes the relation choices, and $p$ refers to the number of choices.
For subject-object pair input, we define the question format $q$ as $q_{\alpha\gamma}$, where $\alpha$ stands for the number of operation terms, and $\gamma$ stands for the modality of the question.
For text-based pair input, the question input $q$ is $q_\text{2text}$, while ${q}_\text{2image}$ is used for the image-based pair input for the two-term subtraction task.
Full prompt examples are provided in the supplementary materials for reference.  
Since this is a classification task, we define the verifiable reward function $R_{2t}$ as~\cite{liu2025visual}:
\begin{align}
    & R_{acc}(a_i)=
\begin{cases}
1& \text{$a_i$=$gt$}\\
0& \text{otherwise}
\end{cases}
\end{align}
\begin{align}
    & R_{2t}(o_i)=R_{acc}(a_i)+R_{format}(o_i)
\end{align}
where ${gt}$ represent the ground truth of the relation and $R_{format}$ is used for ensuring the output of the response $o_i$ match the HTML tags, i.e., $\langle \text{think} \rangle$...$\langle \text{\\think} \rangle$ $\langle \text{answer} \rangle$ $a_i$ $\langle \text{\\answer} \rangle$. $R_{format}$ returns 1 if the format is correct, and 0 otherwise.

\subsubsection{Three-term operation task implementation}
According to the formulation $object_1 - subject_1 + subject_2 = object_2$, we design the prompt below:
\begin{itemize}
    \item Given the relation between $\langle subject_1 \rangle$ and $\langle object_1 \rangle$, please solve the analogy. Answer the following question: $\langle subject_2 \rangle$ is to what as $\langle subject_1 \rangle$ is to $\langle object_1 \rangle$? ...
\end{itemize}
If $\langle subject_1 \rangle$,$\langle object_1 \rangle$,$\langle subject_2 \rangle$ are all replaced by text, we define the question input $q$ as $q_\text{3text}$, while $q_\text{3image}$ for image-based three-term inputs.

Most current verifiable reward functions serve as a binary signal to judge the correctness of the answers, like what we use in two-term subtraction tasks. However, for three-term operation tasks, we are required to assess the semantic similarity between the answer and the ground truth. Therefore, we define the soft verifiable reward function $R_{3t}$:
\begin{align}
    & \mathcal{C}(a,b)=  \frac{\mathbf{E_T}(a) \cdot \mathbf{E_T}(b)}{\|\mathbf{E_T}(a)\| \|\mathbf{E_T}(b)\|}
\end{align}
\begin{align}
    R_{\text{sim}}(a_i) = \sigma \left( \mathcal{C}(a_i,gt) \right) \in (0,1)
\end{align}
\begin{align}
    & R_{3t}(o_i)= R_{sim}(a_i)+R_{format}(o_i) 
\end{align}
where $\mathcal{C}(a, b)$ denotes the cosine similarity between the text embeddings of answer $a_i$ and ground truth $gt$ with text encoders $\mathbf{E_T}$, $gt$ will be replaced by $object_2$, and $\sigma$ is the sigmoid function defined as $\sigma(x) = \frac{1}{1 + e^{-kx}}$.
For the text encoder $\mathbf{E_T}$, we choose Word2Vec instead of a sentence encoder like BERT~\cite{devlin2019bert}, since the subject or object fetched from ConceptNet is a short phrase of up to three words.
Moreover, without the sigmoid function \( \sigma \), the policy model \( \pi_{\theta} \) tends to take shortcuts by generating template responses to gain the reward from $R_{format}$, which is easier than generating the answers to increase the $R_{sim}$, ranging from [-1,1]. 
Therefore, \( \sigma \) ensures that \( R_{\text{sim}} \) is bounded within (0, 1) and amplifies its effect in \( R_{\text{3t}} \).

\section{Experiment}
\subsection{Implementation Details}
\begin{figure*}[t]
    \centering
    \includegraphics[width=\textwidth]{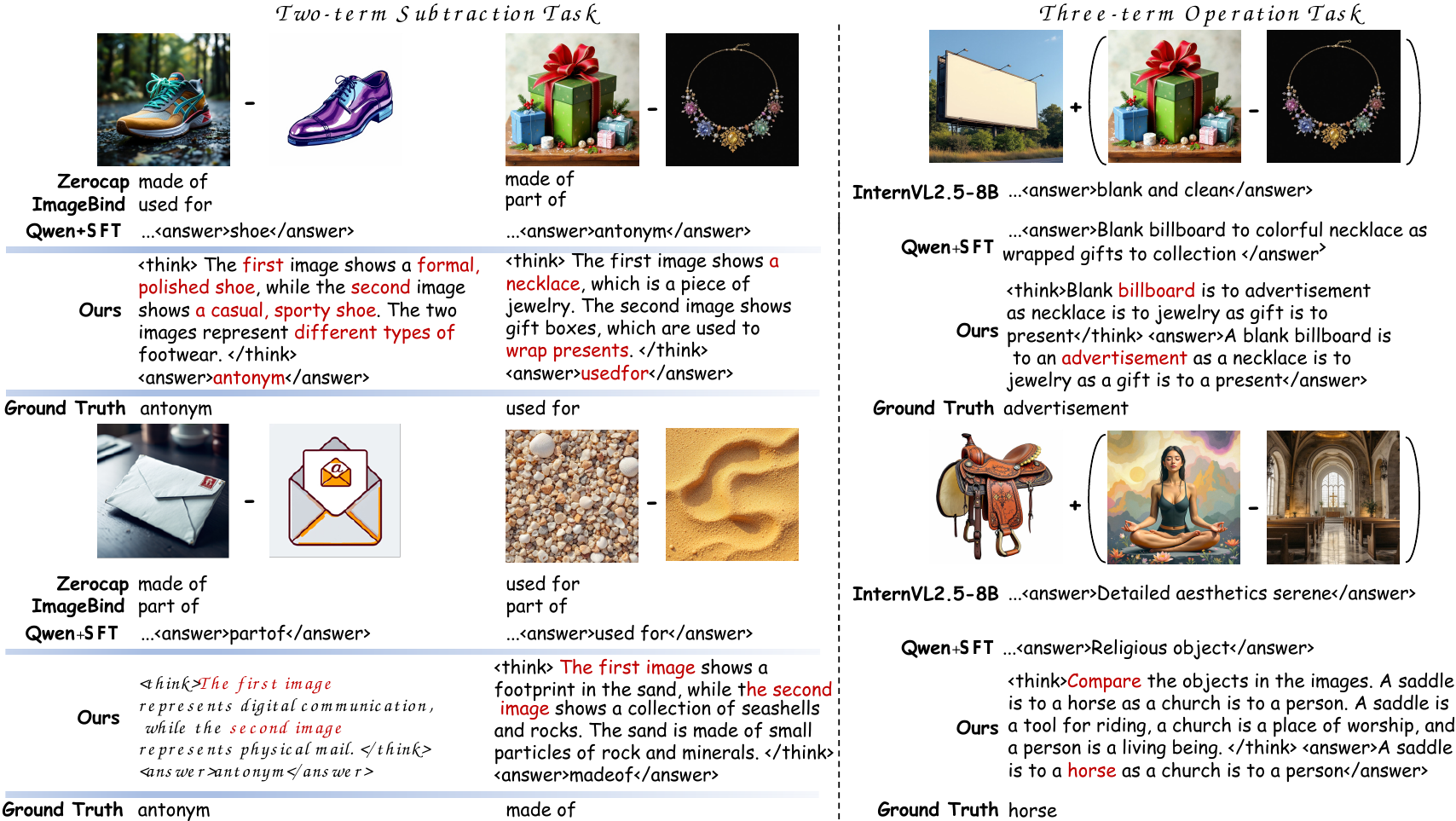}
    \caption{Qualitative results of our model and baselines on the IRPD dataset with image-based question inputs $q_\text{2image}$ and $q_\text{3image}$, where Qwen refers to Qwen2-VL-7B~\cite{bai2023qwen}. Our model shows a clear and coherent thinking process.}
    \label{fig:qualitative_res}
\end{figure*}
\begin{table*}
\centering
\begin{tabularx}{\textwidth}{l *{8}{>{\centering\arraybackslash}X} >{\centering\arraybackslash}X} 
  \toprule[1pt]
  \multirow{4}{*}{Method} & \multicolumn{4}{c}{Two-term Subtraction Task on IRPD} & \multicolumn{4}{c}{Three-term Operation Task on IRPD} & \multirow{4}{*}{\shortstack{Visual7w \\ Telling \\ Accuracy$\uparrow$}} \\
  \cmidrule(lr){2-5} \cmidrule(lr){6-9}
  & \multicolumn{2}{c}{6-option Accuracy$\uparrow$} & \multicolumn{2}{c}{4-option Accuracy$\uparrow$} & \multicolumn{2}{c}{Recall@5 $\uparrow$} & \multicolumn{2}{c}{$\mathcal{C}_\text{word2vec}$$\uparrow$} & \\
  \cmidrule(lr){2-3} \cmidrule(lr){4-5} \cmidrule(lr){6-7} \cmidrule(lr){8-9}
  & $q_{\text{2text}}$ & $q_{\text{2image}}$ & $q_{\text{2text}}$ & $q_{\text{2image}}$ & $q_{\text{3text}}$ & $q_{\text{3image}}$ & $q_{\text{3text}}$ & $q_{\text{3image}}$ & \\
  \midrule
  ZeroCap* & 14.50\% & 18.70\% & 21.14\% & 20.80\% & 0.0074 & 0.0061 & 0.0738 & 0.0652 & - \\
  ImageBind* & 20.99\% & 14.12\% & 29.87\% & 24.49\% & - & - & - & - & - \\
  LanguageBind* & 19.79\% & 13.76\% & 23.28\% & 17.94\% & - & - & - & - & - \\
  \midrule
  InternVL2.5-8B & 49.61\% & 18.70\% & 59.73\% & 28.19\% & 0.0236 & 0.0040 & 0.2169 & 0.1128 & 82.2\% \\
  Video-LLaVa & 30.91\% & 10.30\% & 57.72\% & 24.49\% & 0.0336 & \textbf{0.1010} & 0.2141 & 0.1130 & 38.5\% \\
  Qwen2-VL-7B & 54.96\% & 25.95\% & 64.77\% & 31.54\% & 0.0538 & 0.0741 & 0.2052 & 0.1435 & 82.1\% \\
  Qwen2-VL-7B+SFT & 54.58\% & 23.28\% & 66.11\% & 33.56\% & 0.0522 & 0.0168 & 0.2262 & 0.1521 & 82.1\% \\
  \hline
  Ours & \textbf{58.02\%} & \textbf{35.49\%} & \textbf{66.44\%} & \textbf{51.28\%} & \textbf{0.0781} & 0.0814 & \textbf{0.3135} & \textbf{0.2667} & \textbf{83.8\%} \\
  \bottomrule[1pt]
\end{tabularx}
\caption{Quantitative results on the IRPD dataset and the downstream task on a real-world VQA dataset, Visual7W-telling~\cite{zhu2016visual7w}. Methods marked with * are embedding arithmetic and perform significantly worse than large vision-language models (LVLMs) with reasoning capabilities. Our approach, based on Qwen2-VL-7B and fine-tuned with SAri-RFT, achieves a relative improvement of over 50\% on the two-term subtraction task (4-option) with image-based subject-object pair input $q_\text{2image}$, and over 75\% on the three-term task with $q_\text{3image}$ input, compared to standard supervised fine-tuning (SFT).}
\label{tab:main_result}
\end{table*}

\subsubsection{Dataset Preparation}
We have constructed the IRPD dataset in the format of triplets $relation:subjct \rightarrow object$ across both the text and image domains. While retrieving triplets from Concept’s API~\cite{speer2017conceptnet}, we require each relation to have at least 100 unique subject-object pairs. However, some relations fail to meet this standard through manual verification because they are relatively uncommon, and the triplets retrieved are too abstract to be represented with an image with a low CLIP score.
The dataset example is in Figure ~\ref{fig:flux_dataset}. 

The IRPD dataset cannot be directly used for fine-tuning, so we reconstruct the training and testing datasets in the format of question-answer pairs, where the full prompt templates are proposed in the supplementary material and vary depending on the task type.
For the two-term subtraction task, we constructed two non-overlapping multiple-choice settings to ensure clarity. The 4-option set includes ``usedfor,'' ``antonym,'' ``madeof,'' ``partof,'' while the 6-option set extends this by adding ``atlocation,'' ``createdby.''
The training and testing datasets for the two-term subtraction task each contain 300 question-answer pairs. 
For the three-term operation task, we randomly sample two subject-object pairs, sharing the same relation, to construct the question-answer pairs. The training dataset includes 300 entities, while the testing dataset consists of 1000 entities.

\subsubsection{Experiment Setting}
We report accuracy for two-term subtraction tasks, Recall@5, and cosine similarity using Word2Vec~\cite{church2017word2vec} as text encoder, denoted as $\mathcal{C}_\text{word2vec}$ for the three-term operation task. Recall@5 measures whether any ground-truth word appears among the top five words in the generated response, while $\mathcal{C}_\text{word2vec}$ measures the semantic similarity between the generated answer and ground truth using vector representations.

\subsubsection{Training Setting}

We adopt Qwen2-VL-7B~\cite{bai2023qwen} as our base model, which is an advanced LVLM with 7 billion parameters designed for visual and video understanding, complex reasoning, and multilingual support. Additionally, to accommodate different numbers of input images, the maximum prompt length is set to 1024 for two-term subtraction tasks and 2048 for three-term operation tasks. During the GRPO stage, we set the number $n$ of generations per group as 8. The training stage takes around 2 hours on 8 H800 GPUs with 150 steps. We choose Word2vec as our text embedding to calculate the cosine similarity of the answer and ground truth for the verifiable reward function. 
Moreover, we conducted a user study to evaluate whether Word2Vec similarity scores align with human judgment of the three-term operation task. Specifically, we ranked the model’s outputs using Word2Vec-based similarity scores and selected the top 20\% and bottom 20\% of the results. From each subset, 50 examples were randomly sampled. Five professional annotators rated the reasonableness of the Word2Vec-based similarity score on a 0–5 scale. The average score of 4.2 indicates strong alignment with human preferences.
We also used an LLM for pairwise comparison and showed 79\% agreement with Word2Vec rankings, demonstrating notable consistency.
In contrast, we also tried using CLIP-Score for the text encoder, which led to template-like, uninformative responses, as its consistently high similarity scores fail to reflect true semantic relevance.

\subsection{Results}
\subsubsection{Main Results}
The main results are shown in Table~\ref{tab:main_result}. 
For experiments of ZeroCap~\cite{tewel2022zerocap}, we directly perform the embedding arithmetic of the subject and object to get the answer. For the two-term subtraction task, we calculated the CLIP-score cosine similarity between the model’s generated text output and all relation candidates, selecting the relation with the highest similarity as the final answer.

For multi-modal encoders like ImageBind~\cite{girdhar2023imagebind} and LanguageBind~\cite{zhu2023languagebind}, we subtract the image and text input features at the embedding layer to obtain a vector representing their relation. We then compute the cosine similarity between this vector and the text embedding of each relation, selecting the one with the highest similarity as the final answer. In addition, these two methods cannot be applied to the three-term operation tasks because they cannot generate text outputs with the absence of decoders.

For the experiment of Video-LLava~\cite{lin2023video}, InternVL2.5-8B~\cite{chen2024internvl}, Qwen2-VL-7B~\cite{bai2023qwen}, and Qwen2-VL-7B fine-tuned with Supervised Fine-tuning(SFT), we use the same question prompt template as input to test the model. 

From Table~\ref{tab:main_result}, we observe that our method, SAri-RFT, significantly improves performance on the two-term subtraction task, achieving a 62.59\% gain in 4-option accuracy with the $q_\text{2image}$ input compared to Qwen2-VL-7B fine-tuned via SFT. Moreover, it boosts the $\mathcal{C}_\text{word2vec}$ by 85.85\% with $q_\text{3t,img}$ input and by 52.77\% with $q_\text{3t,txt}$ input, respectively.

\subsubsection{Analysis}
Based on the experimental results, it is evident that embedding arithmetic methods show limited capability in solving semantic arithmetic tasks.
ZeroCap~\cite{tewel2022zerocap} performs poorly primarily due to its training-while-generating paradigm, which is both highly unstable and computationally expensive. 
However, both LanguageBind~\cite{zhu2023languagebind} and ImageBind~\cite{girdhar2023imagebind} perform poorly on the two-term subtraction task, possibly because semantically related text pairs, such as ``woman'' and ``man'', are embedded too closely in the shared space to allow meaningful differentiation.

The qualitative result is shown in Figure~\ref{fig:qualitative_res} with $q_\text{2image}$ and $q_\text{3image}$ inputs. To fairly compare the reasoning capabilities of different LVLMs' reasoning methods, we use identical prompts to elicit their thinking processes during inference, eliminating the influence of prompt variations.
For three-term operation tasks, our model shows the capability of generating answers with analogy formulation: ``A is to B as C is to D,'' which means the model has learned to solve the visual-semantic arithmetic with structured answers.
We present the qualitative results of qualitative result of $q_\text{2text}$ and $q_\text{3text}$ in the supplementary material.
\begin{table}
  \centering
  \begin{tabular}{l cc cc}
    \toprule[1pt]
    \multirow{2}{*}{Method} & \multicolumn{2}{c}{6-option Accuracy$\uparrow$} & \multicolumn{2}{c}{$\mathcal{C}_\text{word2vec}$ $\uparrow$} \\
    & $q_\text{2text}$ & $q_\text{2image}$ & $q_\text{3text}$ & $q_\text{3image}$ \\
    \midrule
    w/ n = 4 & 57.81\% & 33.28\% & 0.3009 & 0.2577 \\
    w/o CoT & 54.21\% & 27.37\% & 0.2231 & 0.0011 \\
    w/o $R_{acc}$ & 53.44\% & 21.37\% & - & - \\
    w/o $R_{sim}$ & - & - & 0.2132 & 0.1614 \\
    w/o $\sigma(\cdot)$ & - & - & 0.2985 & 0.2611 \\
    w/o $R_{format}$ & 53.44\% & 17.56\% & 0.2308 & 0.1399 \\
    \hline
    Ours & \textbf{58.02\%} & \textbf{35.49\%} & \textbf{0.3135} & \textbf{0.2667} \\
    \bottomrule[1pt]
  \end{tabular}
  \caption{Ablation study on two semantic arithmetic tasks. We report accuracy for the two-term subtraction task with six options and $\mathcal{C}_\text{word2vec}$ for the three-term operation task. $n$ denotes the number of generations per group in GRPO, CoT refers to the thinking process when generating the output response. We also show the results of a single reward function without $R_{acc}$, $R_{sim}$, and $R_{format}$, respectively. $\sigma(\cdot)$ is the sigmoid function in the $R_{sim}$.}
  \vspace{-2em}
  \label{tab:ablation}
\end{table}

\subsubsection{Downstream Task on Real-world VQA Dataset}
We evaluate our method alongside other LVLM baselines on a real-world VQA dataset, Visual7W-Telling~\cite{zhu2016visual7w} with 4000 QA pairs, consisting of six different types of questions: what, where, how, who, when, and why. Each question is paired with an image and multiple-choice answers. We adopt our prompt only to fit this scenario, while requiring the models to output both the thinking process and final answer in HTML tags as well. We show the quantitative results on Table~\ref{tab:main_result}. These results demonstrate that our method effectively enhances the model's ability to reason with images.

\subsection{Ablation Study}

In Table~\ref{tab:ablation}, we first demonstrate that increasing $n$, the number of generations per group in GRPO, leads to improved model performance. This suggests that a larger candidate set allows for better selection and optimization.
Next, we explore the effect of modifying the question prompt during response generation. Specifically, we instruct the models to output only the final answer without the thinking process, which causes a dramatic drop in the $\mathcal{C}_\text{word2vec}$ for three-term operation task.
Moreover, we apply both $R_{2t}$ and $R_{3t}$ with format reward $R_{format}$ only during GRPO. In addition, we test our method with a single reward $R_{2t}=R_{acc}$, $R_{3t}=R_{sim}$ for the two-term subtraction task and the three-term operation task, respectively.
The ablation results also show the effectiveness of $\sigma$ used in $R_{sim}$.
\section{Conclusion}

This paper proposes a comprehensive and high-quality IRPD dataset for semantic arithmetic problems, applying GRPO with a newly designed verifiable reward function for the post-training stage, with notable improvement.
However, there are still some limitations. Firstly, the dataset still contains some relatively abstract terms,
such as the presence of abstract terms (e.g., made of: war → violence). 
Moreover, in the RLVR stage, designing a verifiable reward function for the three-term operation task is a significant challenge.
Specifically, during RLVR, the model tends to exploit easier format reward functions
over harder ones,
often leading to only template-like outputs without meaningful content. 
In future work, we plan to explore more arithmetic operations to represent the semantic relations and adopt more methods like latent reasoning to solve these tasks.

\bibliography{aaai2026}
\clearpage
\onecolumn
\section{Appendix}

\subsection{IRPD Dataset}
\begin{figure}[h]
  \centering
  \includegraphics[width=.95\textwidth]{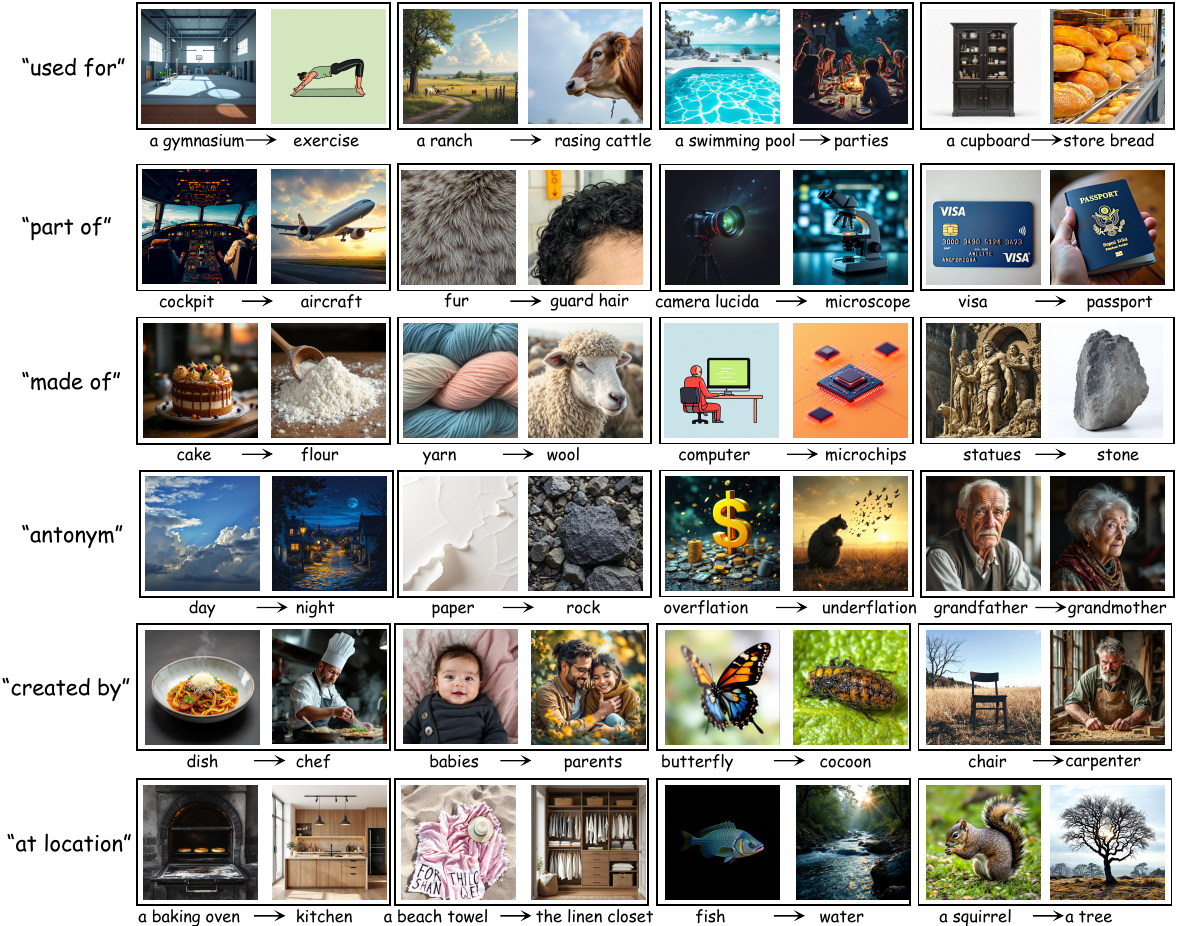}
  \caption{IRPD Dataset Examples. We show 6 relations: used for, part of, made of, antonym, created by, each with 4 examples above.}
  \label{fig:IRPD_dataset_sup}
\end{figure}
Figure~\ref{fig:IRPD_dataset_sup} shows more examples in 6 relations, each with 4 high-quality text-image pairs.

\subsection{Experiment Results}
\subsubsection{Two-term Subtraction Task}
\begin{figure}[h]
  \centering
  \includegraphics[width=0.9\textwidth]{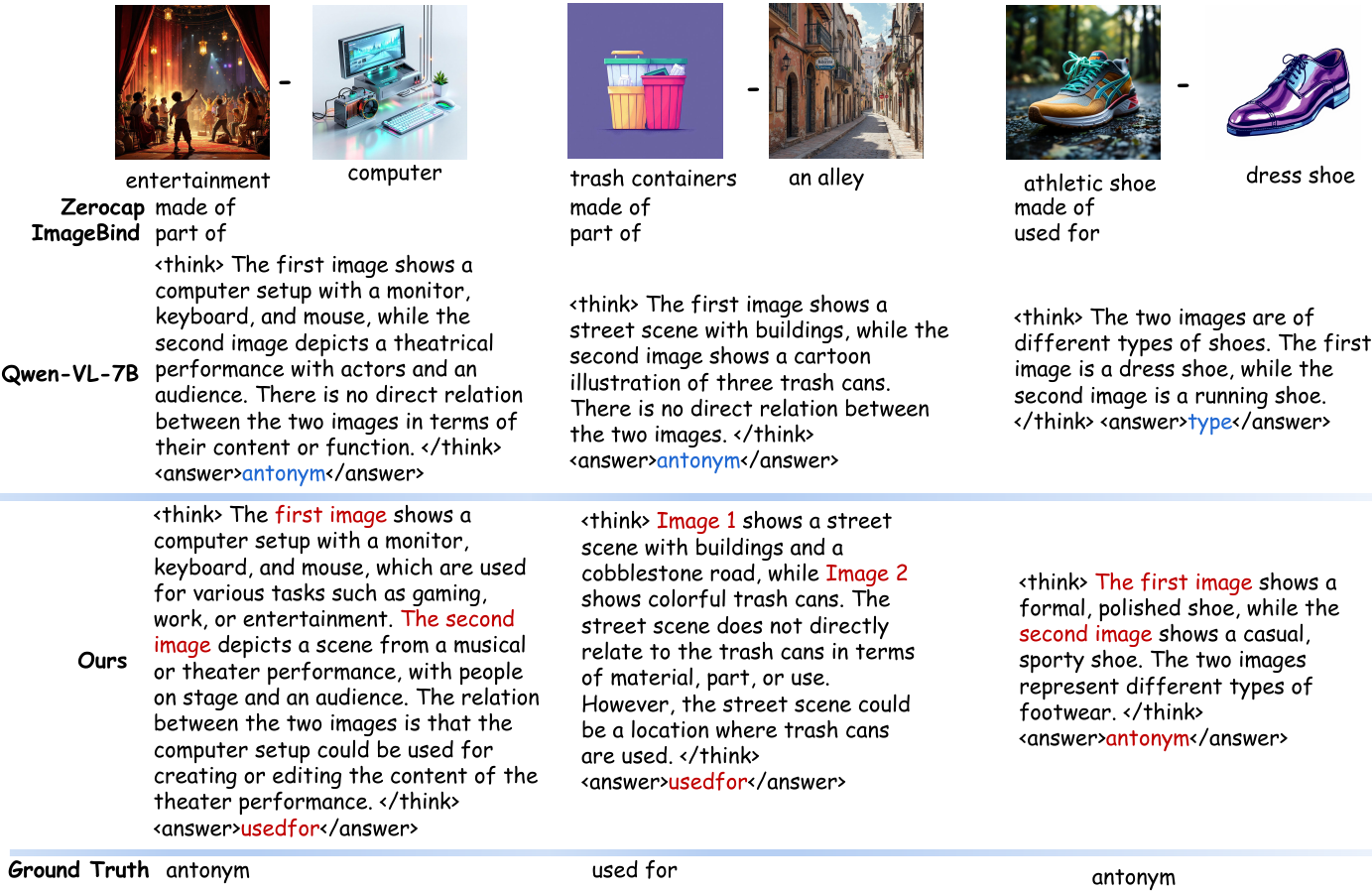}
  \label{fig:sup_1}
\end{figure}

\begin{figure}[h]
  \centering
  \includegraphics[width=0.9\textwidth]{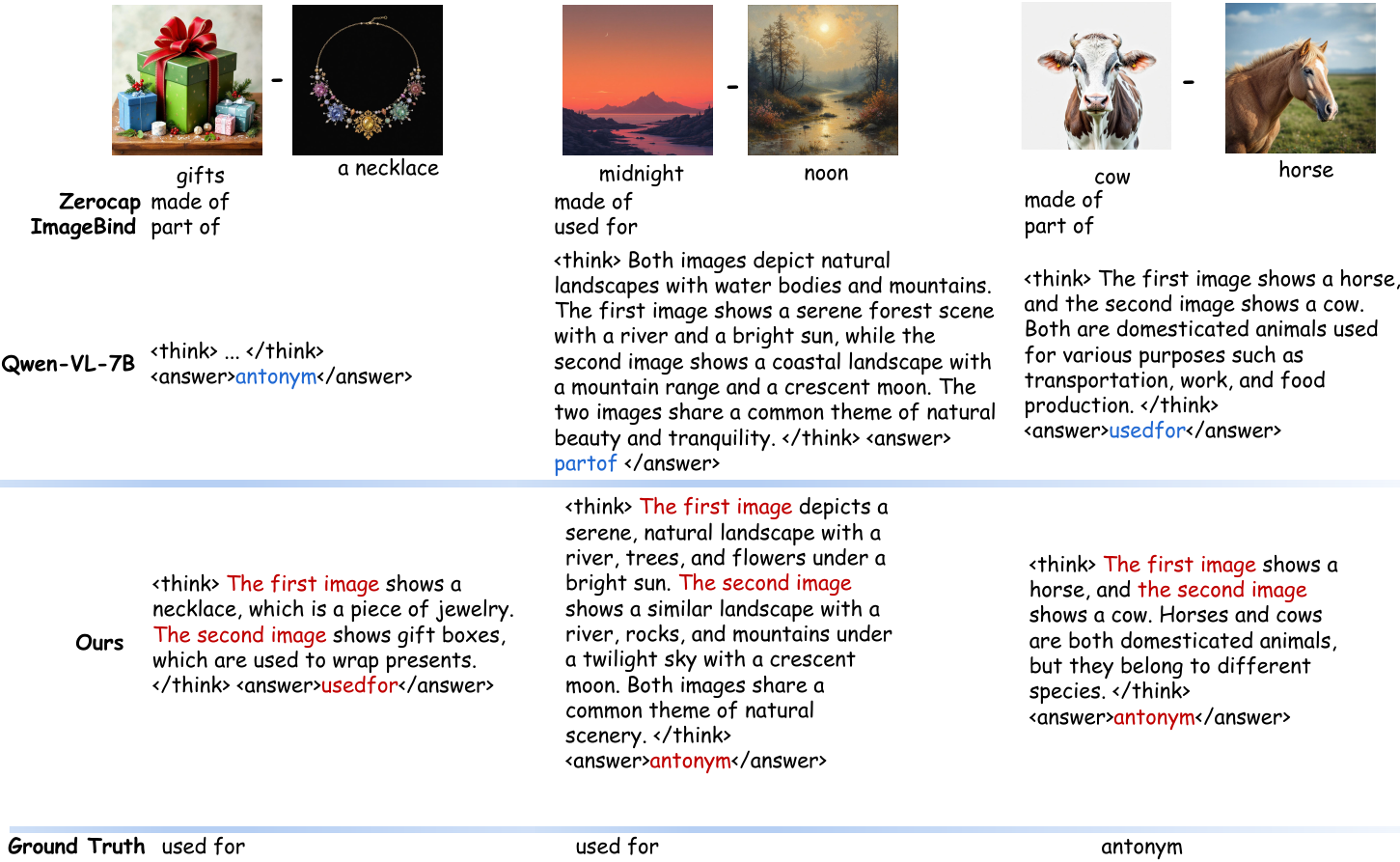}
  \caption{Examples for two-term subtraction tasks, including the comparison of the complete thinking process of Qwen2-VL-7B and our model.}
  \label{fig:sup_2}
\end{figure}
Figure~\ref{fig:sup_2} shows more examples of our two-term subtraction task with image prompts. We show the complete reasoning process of Qwen2-VL-7B and our model.

\subsubsection{Three-term Operation Task}
\begin{figure}[h]
  \centering
  \includegraphics[width=.95\textwidth]{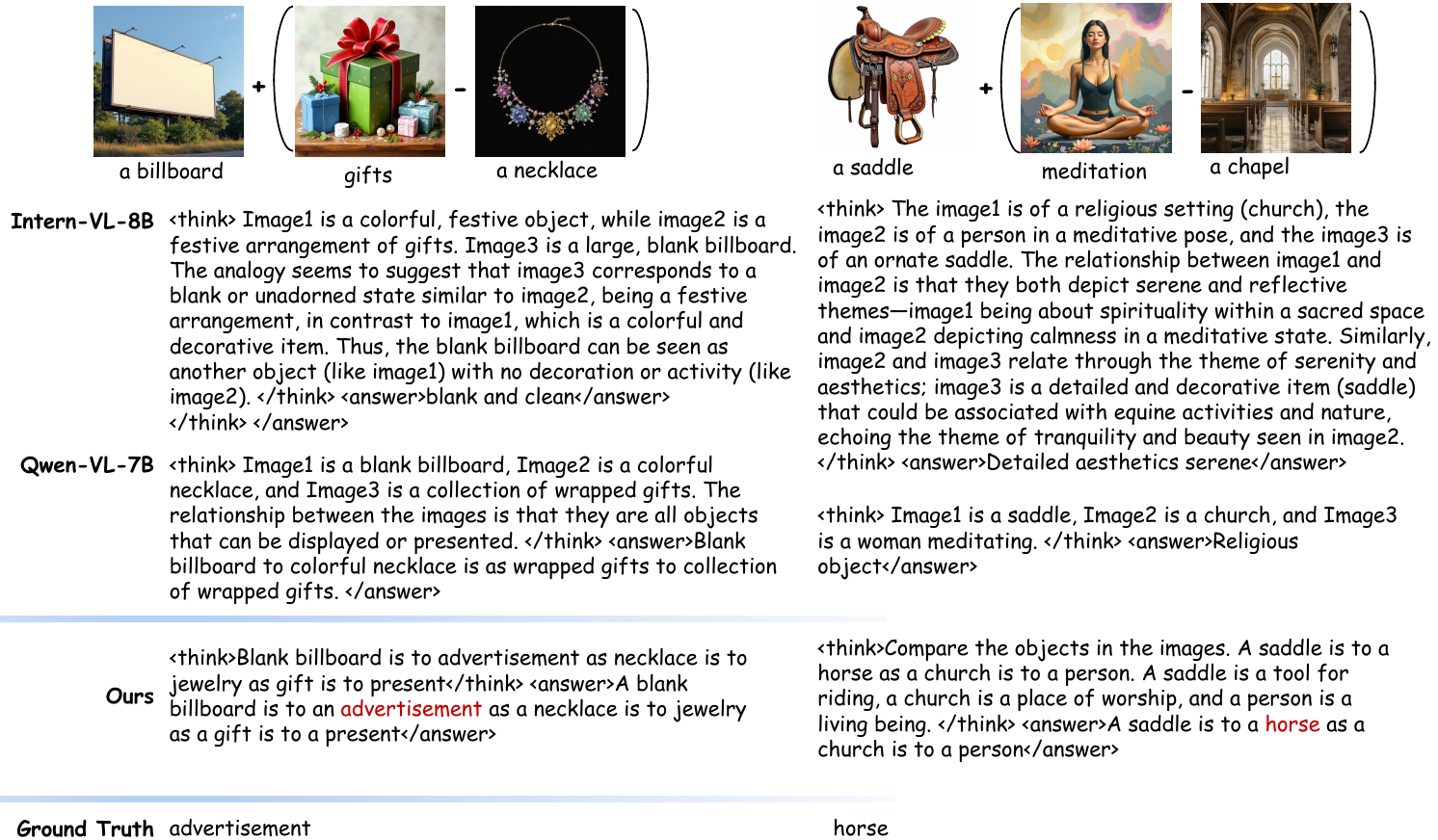}
  \caption{Examples for three-term operation task, including the comparison of the complete thinking process of Qwen2-VL-7B and our model with image prompts.}
  \label{fig:sup_3}
\end{figure}

\begin{figure}[h]
  \centering
  \includegraphics[width=.95\textwidth]{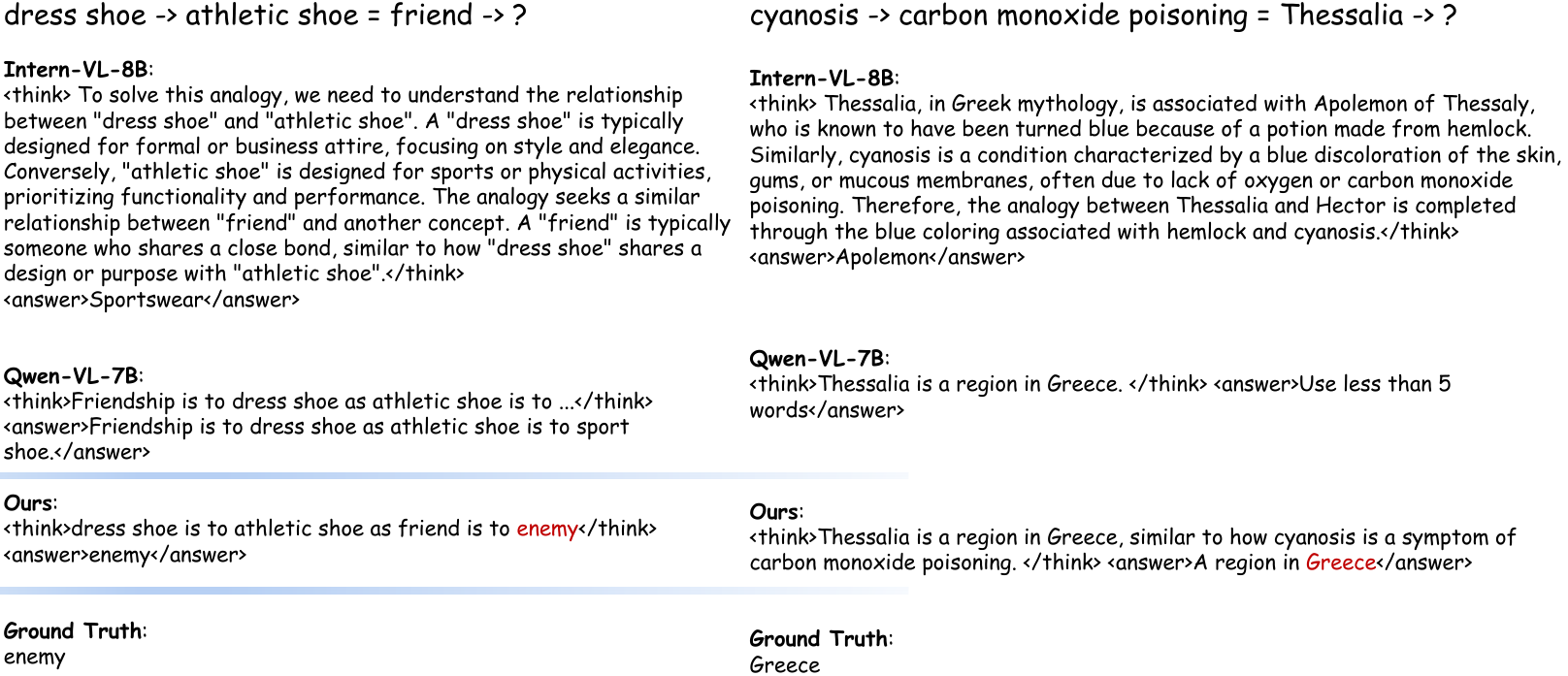}
  \caption{Examples for three-term operation task including the comparison of complete thinking process of Qwen2-VL-7B and our model with text-based question input $q_\text{3text}$.}
  \label{fig:sup_4}
\end{figure}

\begin{figure}[h]
  \centering
  \includegraphics[width=.95\textwidth]{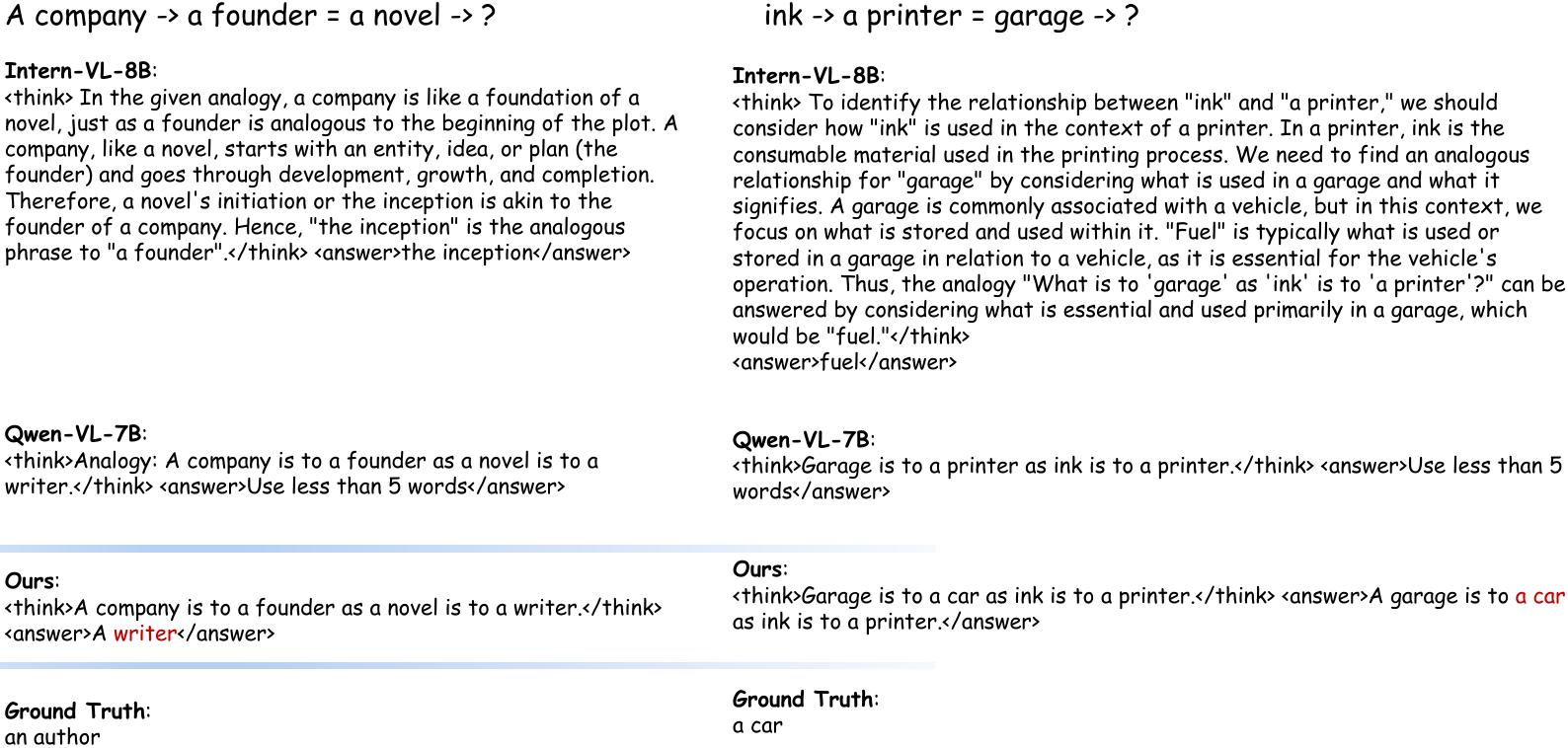}
  \caption{Examples for three-term operation task including the comparison of complete thinking process of Qwen2-VL-7B and our model with text-based question input $q_\text{3text}$.}
  \label{fig:sup_5}
\end{figure}

Figure~\ref{fig:sup_3} shows more examples of our three-term operation tasks with image prompts. 
Figures~\ref{fig:sup_4} and~\ref{fig:sup_5} show more examples of our three-term operation tasks with text prompts.
We show the complete reasoning process of Qwen2-VL-7B and our model for the above two tasks.

\subsection{Prompt Detail}
\begin{figure}[h]
  \centering
  \includegraphics[width=.8\textwidth]{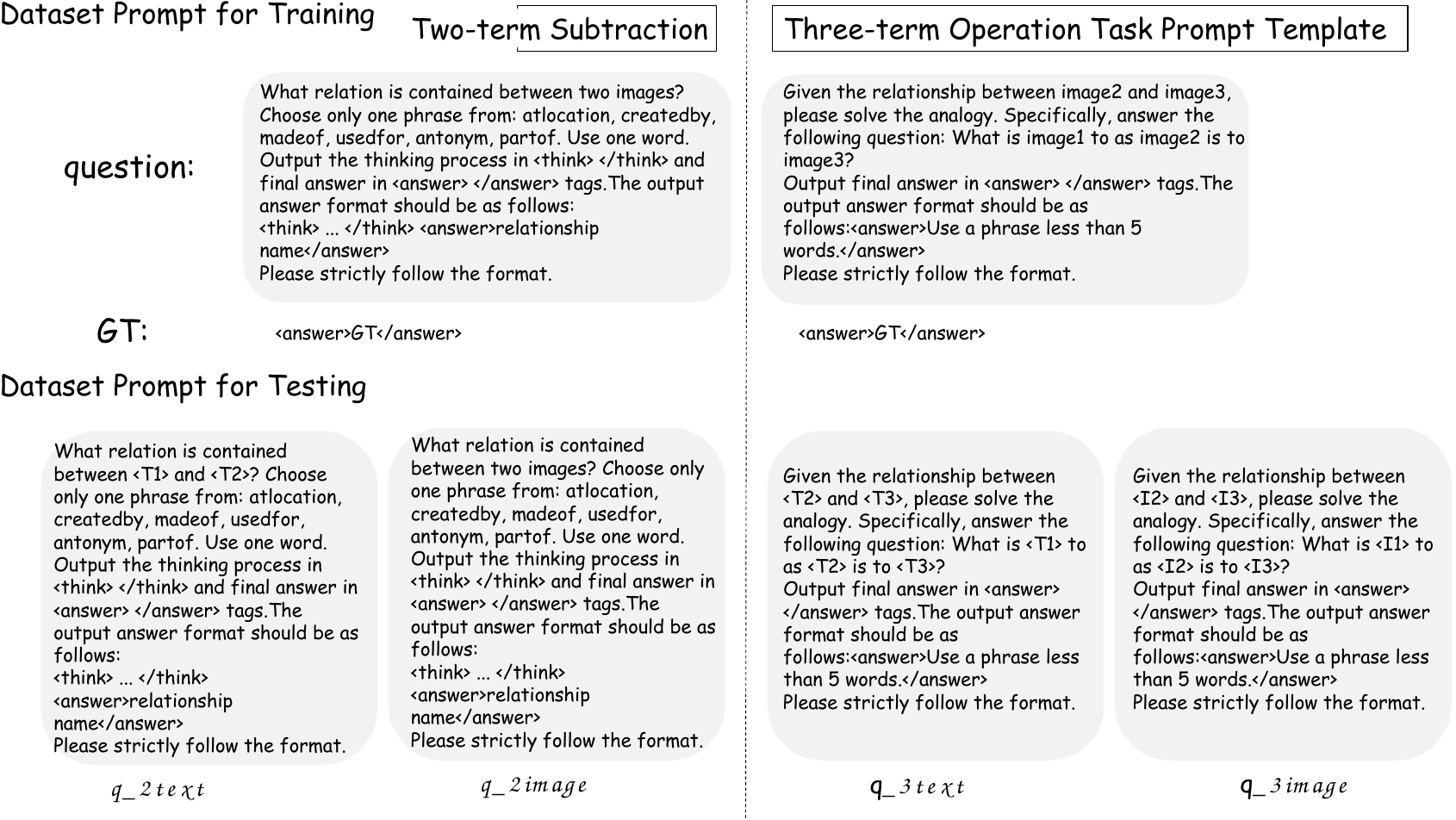}
  \caption{Complete prompt for subtraction and three-term operation tasks for both training and testing stages.}
  \label{fig:sup_6}
\end{figure}
Figure~\ref{fig:sup_6} shows the details of the prompt we constructed the dataset and used for the evaluation stage.

\end{document}